\documentclass[conference]{IEEEtran}
\IEEEoverridecommandlockouts
\usepackage{cite}
\usepackage{amsmath,amssymb,amsfonts}
\usepackage[lined,boxed,commentsnumbered, ruled]{algorithm2e}
\usepackage{algorithmic}
\usepackage{hyperref}
\usepackage{graphicx}
\usepackage{textcomp}
\usepackage{stmaryrd}
\usepackage{xcolor}
\usepackage{amsthm}
\usepackage{float}
\usepackage{subfig}
\usepackage{makecell}
\usepackage{booktabs}
\usepackage{caption}
\usepackage{fancyhdr}
\usepackage{multirow}
\usepackage{authblk}

\usepackage{float}
\usepackage{subfig}
\usepackage{makecell}

\newcommand{\encrypt}[1]{{\llbracket #1 \rrbracket}}
\newcommand*{\affaddr}[1]{#1} % No op here. Customize it for different styles.
\newcommand*{\affmark}[1][*]{\textsuperscript{#1}}
\newcommand*{\email}[1]{\texttt{#1}}
\newtheorem{prop}{Proposition}
\def\BibTeX{{\rm B\kern-.05em{\sc i\kern-.025em b}\kern-.08em
    T\kern-.1667em\lower.7ex\hbox{E}\kern-.125emX}}
\begin{document}

\title{Fed-EINI: An Efficient and Interpretable Inference Framework for Decision Tree Ensembles in Vertical Federated Learning
}

\author{Xiaolin~Chen\affmark[1,2], Shuai~Zhou\affmark[4], Kai~Yang\affmark[3], Hao~Fao\affmark[3], Hu.~Wang\affmark[3] and Yongji.~Wang\affmark[1.2]\\
\affaddr{\affmark[1]Cooperative Innovation Center, Institute of Software, Chinese Academy of Sciences, Beijing, China} \\
\affaddr{\affmark[2]University of Chinese Academy of Sciences, Institute of Software, China};\\
\affaddr{\affmark[3]JD Technology Group, Beijing, China};\\
\affaddr{\affmark[4]University of Technology Sydney, Australia};\\
\email{wanghu5@jd.com}\\
\email{ywang@itechs.iscas.ac.cn}\\
}

\maketitle

\begin{abstract}
  The increasing concerns about data privacy and security drive an emerging field of studying privacy-preserving machine learning from isolated data sources, i.e., \textit{federated learning}.
  A class of federated learning, \textit{vertical federated learning}, where different parties hold different features for common users, has a great potential of driving a great variety of business cooperation among enterprises in many fields. In machine learning, decision tree ensembles such as gradient boosting decision trees (GBDT) and random forest are widely applied powerful models with high interpretability and modeling efficiency. However, state-of-art vertical federated learning frameworks adapt anonymous features to avoid possible data breaches, makes the interpretability of the model compromised.

  To address this issue in the inference process, in this paper, we firstly make a problem analysis about the necessity of disclosure meanings of feature to Guest Party in vertical federated learning. Then we find the prediction result of a tree could be expressed as the intersection of results of sub-models of the tree held by all parties. With this key observation, we protect data privacy and allow the disclosure of feature meaning by concealing decision paths and adapt a communication-efficient secure computation method for inference outputs. The advantages of Fed-EINI will be demonstrated through both theoretical analysis and extensive numerical results. We improve the interpretability of the model by disclosing the meaning of features while ensuring efficiency and accuracy.
\end{abstract}


\section{Introduction}
Data in the real world widely exist in isolated islands, which becomes a fundamental limiting factor for AI modeling and data analytics. For example, personal credit assessing in data-driven risk management usually uses three types of data: qualification data, credit data, consumption data, often held by different enterprises and institutions. The security and privacy of data are increasingly concerned in recent years. Typically, operators have enacted laws and regulations to protect the privacy of users' data, such as the General Data Protection Regulation (GDPR) \cite{EU} by European Union and California Consumer Privacy Act by California, United States. Moreover, the data assets are highly valued, making data owners often reluctant to reveal their data to others. It drives the emergence of a novel field, termed as \textit{federated learning} \cite{konevcny2016federated,yang2019federated,kairouz2019advances}, studying privacy-preserving distributed machine learning from multiple data sources without sharing original data.

\textbf{Backgrounds and Related Work.} According to the structure of sample space and feature space across data sources, federated learning can be categorized into three classes \cite{yang2019federated}, i.e., horizontal federated learning, vertical federated learning, and federated transfer learning. Horizontal federated learning \cite{konevcny2016federated,hamer2020fedboost} refers to the scenarios where each participating party holds a subset of all data samples with common feature space. Vertical federated learning \cite{yang2019quasi,gu2020federated} studies collaborative machine learning where different parties share the same sample space but differ in feature space. Federated transfer learning \cite{liu2020secure} usually explores the area where different parties differ both in sample space and feature space. Among these structures, vertical federated learning is a promising approach to bridge the gap between isolated data providers for business cooperations and beyond the limits of locally available data to AI systems. There are a line of works studying the vertical federated learning of linear/logistic regression \cite{mohri2019agnostic,yang2019quasi}, gradient boosting decision tree (GBDT) \cite{cheng2021SecureBoost}, random forest \cite{liu2020federated}, kernel methods \cite{gu2020federated}, etc.

  Decision tree, especially decision tree ensemble models \cite{zhou2009ensemble} including random forest and GBDT \cite{friedman2001greedy}, are an important class of machine learning models due to their powerful generalization capability, high modeling efficiency, and better interpretability. Federated learning of decision tree models \cite{cheng2021SecureBoost,feng2019securegbm,liu2020federated,fang2008privacy,ong2020adaptive} has garnered numerous attention recently. The works of \cite{li2020practical,ong2020adaptive} focused on improving the efficiency of GBDT model training of horizontal federated learning. SecureBoost \cite{cheng2021SecureBoost} first proposed a vertical federated learning framework to build GBDT model, whose security is guaranteed via encrypting the exchanged intermediate values with homomorphic encryption (HE). SecureGBM \cite{feng2019securegbm} proposed to extend the efficient GBDT framework LightGBM to vertical federated learning, and \cite{liu2020federated} proposed a federated random forest framework.

  Although vertical federated decision tree ensemble models have been widely applied in many fields, unfortunately, we observe that the inference of existing works such as SecureBoost and secureGBM severely compromise the interpretability of the model due to anonymous features. In these works \cite{cheng2021SecureBoost,feng2019securegbm,liu2020federated}, the commonly used multi-interactive (MI) inference procedure makes decision path of Host parties (i.e., data providers) known to the Guest party (i.e., the model user), which brings possible data breaches of Host with disclosed feature meaning. The federated learning models without explicit feature meanings are not interpretable in business applications. Both GDPR in EU and the Equal Credit Opportunity Act (ECOA) \cite{hsia1978credit} in US grant users the "right to explanation". A new financial industry standard \cite{JR/T0221-2021} recently released by the People’s Bank of China specifies the interpretable requirement for features used in the AI algorithms for financial applications.

  \textbf{Our Work.} The state-of-art vertical decision tree ensembles adapt a MI(multi-interactive) inference with the anonymous feature for possible data breaches. In financial applications, since the model's interpretability is critical for the user, we focus on improving the interpretability of the vertical decision tree ensembles. To address the issue, we shall propose Fed-EINI, a secure and interpretable inference framework allowing to disclose feature meanings of Host parties to the Guest party. We observe that possible data breaches of existing works result from unprotected decision paths, which motivates the adoption of HE to compute the inference result with encrypted decision paths securely. To improve the communication efficiency and achieve low-latency inference, we calculate, encode and encrypt candidate decision paths locally at each party to make the decision path indistinguishable by other parties. The inference result is then obtained via secure aggregation with only one round of encrypted information exchange. We improve the interpretability of the model by disclosing the meaning of features while ensuring efficiency and accuracy.
%   We focus on improving the interpretability of federated decision tree ensembles for practice bussiness. In many application, after building the model, the model should be explainable to the Guest party. To address the issue, in this paper, we shall propose Fed-Eini, a secure and interpretable inference framework allowing to disclose feature meanings of Host parties to the Guest party. We observe that possible data breaches of existing works result from unprotected decision paths, which motivates the adoption of HE to compute the inference result with encrypted decision paths securely.To improve the communication efficiency and achieve low-latency inference, we compute, encode and encrypt candidate decision paths locally at each party to make the decision path indistinguishable by other parties. The inference result is then obtained via secure aggregation with only one round of encrypted information exchange.

  \textbf{Our Contributions.} We summarize the contribution of the proposed Fed-EINI framework in following aspects.
  \begin{itemize}
  \item Firstly, we explained the necessity of feature disclosure from the perspective of industry compliance and model users, which is the bottleneck in the application of vertical federated learning in the financial field. The existing works, such as SecureBoost, do not disclose the meaning of the features for data privacy. In Fed-EINI we proposed, based on a key observation, the Host party discloses the meaning of features to the Guest party for interpretability and encrypts the decision path for data privacy.
  \item Secondly, in terms of security, we compared the information available to each party in Fed-EINI with the existing vertical federated decision tree ensembles. Our method allows the disclosure of the meanings of features held by passive parties to the Guest while hiding the splitting path information. Therefore, the detailed privacy of each sample will not be leaked to the Guest party, and the splitting path in Fed-EINI is secured by Paillier cryptosystem.
  \item Finally, we use theoretical analysis and numerical results conducted to demonstrate the efficiency and accuracy of Fed-EIN. As the prediction result of a single tree can be expressed losslessly as the intersection of results of the sub-models of the trees held by all parties. Fed-EINI can achieve the same accuracy as the existing method. Due to the adoption of a more efficient communication-efficiency calculation method, we decouple the calculation process between multiple parties, and the encoding process of each party does not depend on the results of other parties. There will be less waiting time and communication rounds between multiple parties. The whole inference time is reduced by more than 38\% on the experiment result.
  \end{itemize}

The rest of the paper is organized as follows. In Section II, we introduce federated decision tree ensemble models and the inference problem, and then we discuss the problem of laking interpretability. In Section III, there is a two-stage framework, Fed-EINI. In Section IV, we analyze the proposed algorithms on interpretability, security, accuracy, and efficiency. In Section V, we evaluate Fed-EINI using different scale datasets with baseline SecureBoost. Finally, we conclude the paper in Section VI.

\section{Preliminaries and Problem Definition}
This section introduces the vertical federated decision tree ensemble models and states the federated inference problem. We then summarize the multi-interactive (MI) inference framework adopted in state-of-the-art works and discuss the challenging problem of lacking interpretability.Table \ref{summary} summarizes the frequently used notations.
\begin{table}[htb]
\centering
\caption{Summary of notations}
\label{summary}
\begin{tabular}{  l  l }
\toprule
Notation & Description \\
\toprule
$N$&  the number of samples in the prediction dataset \\
$d$ & the number of total features \\
$M$ & the number of parties in vertical FL \\
$K$ & the number of total trees \\
$\mathbf{x}^m$ & the features held by the m-th part  \\
$P^m$ & the m-th party  \\
$f_k$ & the k-th decision tree \\
$f_k^m$& the sub-model of k-th decision tree held by m-th party \\
$Guest$ &the model user  as well as the active party\\
$Host$ &the passive party\\
$\encrypt{W_k^{Guest}}$ &the Guest party's encrypt decision vector\\
$W_k^{Host}$ &the Host party's decision vector\\
$S_k$ &the prediction of sample in the k-th tree \\
$\hat{y}$ & the prediction of sample in the classification task \\
$w_{(j,k)}$ &the weight of the j-th leaf of the k-tree\\
\bottomrule
\end{tabular}
\label{table:notations}
\end{table}

\subsection{Vertical Federated Decision Tree Ensemble Models}\label{subsec:problem_statement_model}
Consider a vertical federated learning system with $M$ parties denoted by $P^m,m \in \{1,2,\cdots,M\}$. Denote the datasets distributed in $M$ parties as $\{\mathbf{X}^m\}_{m=1}^M$. The local dataset of party $m$, i.e., $\mathbf{X}^m\in \mathbf{R}^{n \times d_m}$, consist of $d_m$ features and $n$ data samples. Therefore, the datasets of $M$ parties can be looked as a vertically split on the large dataset $\mathbf{X}=[\mathbf{X}^1,\cdots,\mathbf{X}^M]\in\mathbf{R}^{n \times d}$ with disjoint feature space, in which $d\!=\!\sum_{m=1}^{M}d_m$. Vertical federated learning is closely concerned with business cooperation where each party is usually a different company, and the data labels are available by only a single party called the Guest party. Other participants without labels are called Host parties. The Guest party performs as the model user requiring more data features from Host parties to improve the performance of the AI model.

Here we take SecureBoost as a representative example to show how to build a federated decision tree ensemble model interactively. SecureBoost is a lossless extension of XGBoost \cite{chen2016xgboost}, an efficient GBDT modeling framework, to vertical federated learning. We shall greedily build $K$ regression trees $\{f_k\}_{k=1}^{K}$, where the $t$-th tree model $f_t$ is built by minimizing the second-order approximation loss function
\begin{equation}\small
    L^{(t)} \simeq \sum_{i=1}^N \Big(l(y_i,\hat{y}_i^{(t-1)})+g_if_t(x_i)+\frac{1}{2}h_if_t^2(x_i)\Big)+\Omega(f_t),
\end{equation}
where $l(y_i,\hat{y}_i)$ defines the loss between prediction value $\hat{y}_i$ and target label $y_i$, $g_i = \partial_{\hat{y}^{(t-1)}} l(y_i,\hat{y}^{(t-1)})$ is the gradient value, $h_i = \partial^2_{\hat{y}^{t-1}}l(y_i,\hat{y}^{t-1})$ is the Hessian value, $\hat{y}_i^{(t-1)}=\sum_{k=1}^{t-1}f_k(x_i)$ is the prediction result of previous $t-1$ trees, and $\Omega(f_t) = \gamma T+\frac{1}{2}\lambda||\mathbf{w}||^2$ is the regularization term with $\gamma>0,\lambda>0$. $T$ is the number of leaves and $\mathbf{w}$ is the vector of weights on leaves.

The key procedure is node generation by interactively computing the information gain for each feature and each possible splitting rule and computing the weights of leaf nodes. Both information gain and weights are functions of the aggregated gradients and Hessians, which can only be computed by the Guest party based on labels. Therefore, SecureBoost proposes that Guest encrypts all gradient $g_i$ and Hessian $h_i$ with additively homomorphic encryption such as Paillier \cite{paillier1999public} and transmits them to each Host party. Host parties compute the corresponding aggregation results of encrypted gradient and Hessian, and transmit them back to the Guest party. The Guest party can then compute the information gain for all possible splitting points, as well as the weights if the node is a leaf node.

With additively homomorphic encryption \cite{paillier1999public} denoted by $\encrypt{\cdot}$, we can efficiently compute the sum and scalar multiplication with ciphertexts, i.e., $\encrypt{u}+\encrypt{v}=\encrypt{u+v}$ and $v\cdot\encrypt{u}=\encrypt{v\cdot u}$.

\subsection{Vertical Federated Inference Problem}
In this subsection, we provide the mathematical description of the inference problem for federated decision tree ensemble models. In the inference process, multiple parties have already collaboratively trained $K$ decision tree models $\{f_k\}_{k=1}^K$. Let the input data sample for inference be $\mathbf{x}=(x_1,x_2,...,x_d)=(\mathbf{x}^{1},\mathbf{x}^{2},...\mathbf{x}^{M})\in \mathbf{R}^{d}$, where $\mathbf{x}^{m}\in \mathbf{R}^{d_m}$ is the feature uniquely held by party $m$.

Given the trained decision tree ensemble model $\{f_k\}_{k=1}^K$ and input sample $\mathbf{x}$, the inference result can be expressed as
\begin{equation}
\hat{y}=\mathcal{G}(f_1(\mathbf{x}),f_2(\mathbf{x}),...,f_K(\mathbf{x})),
\end{equation}
where $\mathcal{G}$ is the ensemble strategy of prediction values of all decision tree models. In vertical federated learning, each participant holds a subset of all nodes in each $f_k$, and the leaf weights are typically available only by the Guest party. The inference results should be computed securely without sharing feature data among parties.

In the following, we present the expressions of two typical ensemble models, gradient boosting decision tree and random forest.
\begin{itemize}
\item \textbf{GBDT:} As presented in Section \ref{subsec:problem_statement_model}, the inference result of GBDT model is given by the sum of $K$ trees' prediction values, i.e.,
\begin{equation}
\hat{y}=\sum\limits_{k=1}^Kf_k(\mathbf{x})=\sum\limits_{k=1}^K\sum\limits_{j\in T_k}w_{(j,k)}\mathbf{I}(\mathbf{x}\in Leaf_j),
\end{equation}
where $T_k$ is the index set of all leaf nodes for the $k$-th tree, $x\in Leaf_j$ represents that the prediction result of data x is the weight of $j$-th leaf node.
\item \textbf{Random Forest:} Random forest is a well-known bagging tree method by independently learning many decision trees on a randomly sampled subset of data samples and features. The inference result is given by averaging the prediction values of $K$ trees, i.e.,
\begin{equation}\small
  \hat{y}=\frac{1}{K}\sum\limits_{k=1}^Kf_k(\mathbf{x})=\frac{1}{K}\sum\limits_{k=1}^K\sum\limits_{j\in T_k}w_{(j,k)}\mathbf{I}(\mathbf{x}\in Leaf_j).
\end{equation}
\end{itemize}

\subsection{Multi-Interactive (MI) Inference}
\begin{figure}[tb]
\centering
\centerline{\includegraphics[width=\columnwidth]{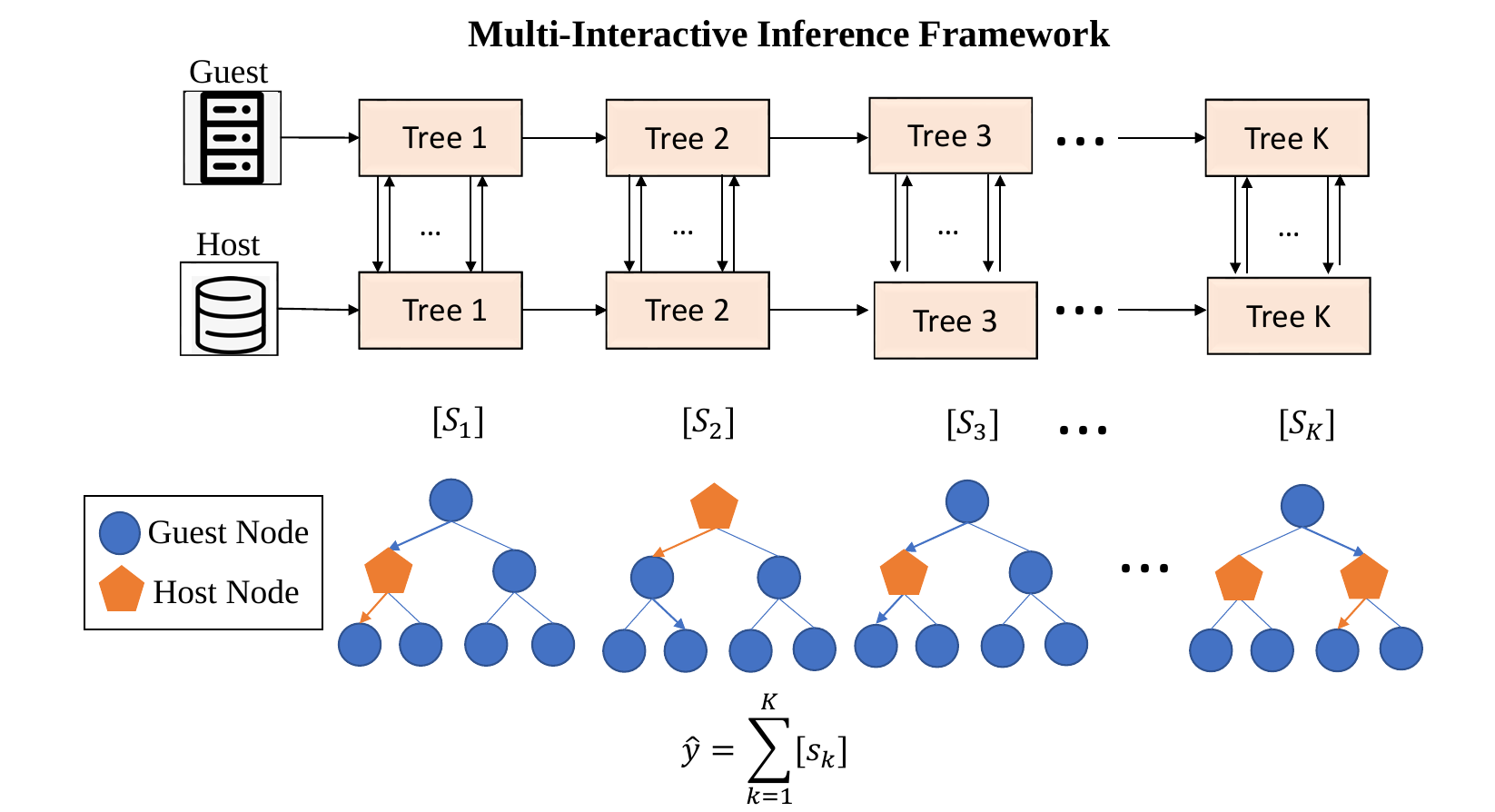}}
\caption{Structure of interactions for the multi-interactive inference framework.}
\label{fig:interaction_structure}
\end{figure}

As shown in Figure \ref{fig:interaction_structure}, existing works \cite{cheng2021SecureBoost,feng2019securegbm} have adopted a MI inference framework to compute the inference result. In this framework, the decision path of a decision tree model is determined by sequentially finding the decision on the current node and moving to the next node. Each decision is made by evoking the party which owns the corresponding feature and sending the computation result back to the Guest. If the splitting rule of the node and the feature are owned by the Guest party, the Guest party makes the decision path based on local data. If the node and the feature are owned by others, the Guest party should receive the decision path from the Host party which owns the corresponding feature. The weight on the leaf node reached by the decision path is the decision tree's prediction value. The MI inference framework admits a serial structure of interactions shown in Figure \ref{fig:interaction_structure}, which means the inference is performed tree by tree. Finally, the Guest party summarizes all the weights for prediction. Note that the features for Host parties should be anonymous to the Guest, and the Guest party has only the index numbers (e.g., $HF_1$, $HF_2$) to avoid possible data breaches.
\begin{figure}[h]
\centering
\centerline{\includegraphics[width=\columnwidth]{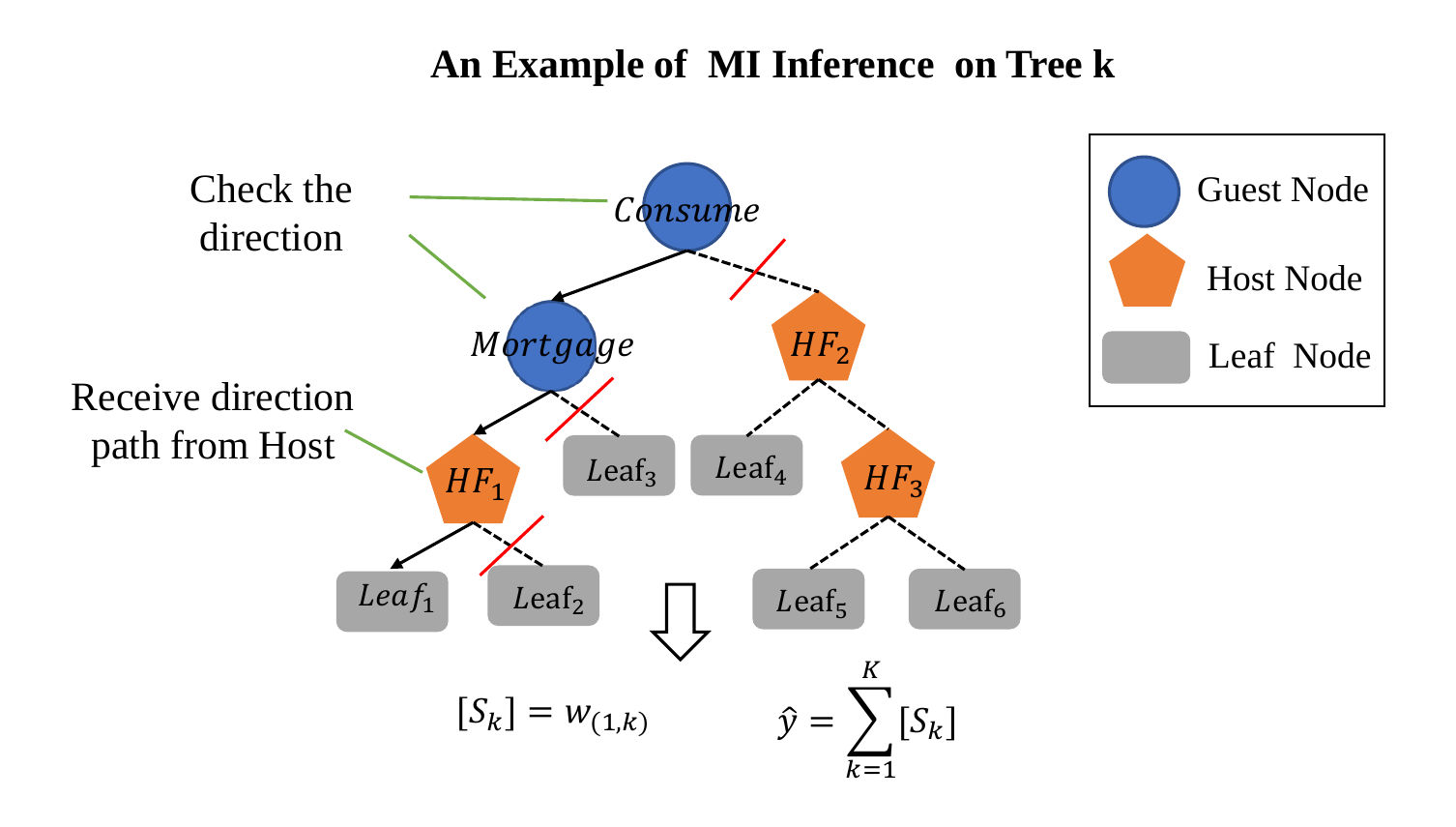}}
\caption{An example for MI inference}
\label{fig:interaction_example}
\end{figure}

Here we show a detailed example of the MI inference on the k-th tree as illustrated in Figure \ref{fig:interaction_example}. The Guest party knows the local splitting rules of blue nodes named ``Consume'' and ``Mortgage'' but do not know the rules of orange nodes named $HF_1,HF_2,HF_3$. For sample $\mathbf{x}$, the Guest party makes the decision on its own blue nodes with local splitting rules, and then it receives the decision path on the node named $HF_1$ from the Host party. The weight of $Leaf_1$ is the inference result $S_k$ of the k-th tree for $\mathbf{x}$. Finally, the Guest party summarizes all the inference results of trees for prediction.
\subsection{Problem Analysis}\label{subsec:problem_analysis}
The pioneering MI inference framework enables a party to make predictions enhanced by external data sources. However, the features for Host parties should be anonymous to Guest and the Guest party has only the index numbers for the MI inference framework in existing works \cite{cheng2021SecureBoost,feng2019securegbm}.  Disclosing the meanings of features of Host parties to Guest leads to possible data breaches. For example, if a feature in Host indicates whether a user ``is in blacklist", we know that bad guys usually take only a small portion of all users. Therefore, the splitting direction with fewer samples indicates that the corresponding users are in the blacklist while others are not.

Model interpretability is urgently needed to realize the commercial value of federated learning in many fields such as financial risk management and smart medical diagnosis and treatment. Both GDPR \cite{EU} enacted by EU and ECOA \cite{hsia1978credit} in US grants users the ``right to explanation". Recently, the People's Bank of China released the financial industry standard titled \textit{evaluation specification of AI algorithm in financial application} \cite{JR/T0221-2021}, which specifies the interpretable requirement for features. The meanings of features are critical and mandatory for model interpretability. There are mainly two reasons as listed below.
\begin{itemize}
\item Firstly, the meanings of features are essential to prove the compliance of the model to regulators in financial businesses. For example, consumers should not be discriminated by their gender, age, race, and religion as required by ECOA in US. As shown in Figure \ref{fig:logo}, a credit risk management model with the gender feature is not allowed, and gender information should be removed from the model. This problem cannot be addressed with the existing MI inference framework in SecureBoost because of the anonymous features.
\item Moreover, industry practitioners need the meanings of features to judge the reasonability of AI models, especially for financial risk management. Since financial risks are usually impossible to be quantified explicitly, financial industry practitioners are often careful in making decisions to avoid potential loss. They usually evaluate AI models with their professional knowledge and experience before use, which requires the meanings of used features. For example, according to The expert experience, people with more stable consumption behavior should have higher credit scores. AI models behaving the opposite way are thought to be unreasonable and unusable for financial industry practitioners.
\end{itemize}%\vspace{-0.5em}

To provide interpretability for inference of vertical federated decision tree ensemble models, we propose to encrypt the decision path to guarantee the security of disclosing the meanings of Host parties' features. However, directly encrypt the decision path following MI inference framework requires fully homomorphic encryption methods\cite{van2010fully} supporting both addition and multiplication, which brings unaffordable computation overhead in applications. To address these challenges, in this paper, we shall propose an efficient and interpretable inference framework named Fed-EINI, which adopts an additively homomorphic encryption method to secure decision paths.
% However, encrypting the decision direction layer by layer following the MI working flow will result in high communication overhead due to multiple rounds of exchanging ciphertexts. In this paper, we shall propose an efficient and interpretable inference framework, which securely computes the inference result via only one round of ciphertext exchange without disclosing the decision path to the Guest party.
% \begin{figure}[tb]
% \vskip 0.2in
% \centering
% \centerline{\includegraphics[width=\columnwidth]{secureboost1_1.png}}
% \caption{Structure of interactions for the multi-interactive inference framework.}
% \label{fig:interaction_structure}
% \vskip -0.2in
% \end{figure}
% \begin{figure}[tb]
% \vskip 0.2in
% \centering
% \centerline{\includegraphics[width=\columnwidth]{secureboost_1.png}}
% \caption{Structure of interactions for the multi-interactive inference framework.}
% \label{fig:interaction_structure}
% \vskip -0.2in
% \end{figure}
\section{Fed-EINI: Proposed Two-Stage Framework}
This section presents Fed-EINI, a two-stage inference framework adopting additively homomorphic encryption to secure the decision path, including \textit{parallel calculation} stage and \textit{synchronization} stage.

\subsection{Fed-EINI Algorithm}
A tree model maps $d$ features $(x_1,\cdots,x_d)$ of a sample $\mathbf{x}$ to the weight of a certain leaf node. A decision path is actually a combination of rules defined by the nodes on the path. Therefore, the output of a decision tree model $f(\mathbf{x})$ is the weight of the leaf node whose corresponding path from the root node to the leaf node holds conditions that can be simultaneously satisfied by the input data sample $\mathbf{x}$.

Denote the sub-model party $m$ held for the $k$-th tree $f_k$ as $f_k^{m}$. The sub-model is defined by a combination of rules according to the locally owned nodes of party $m$.
Any party $m$ can verify if a leaf node is a possible inference output based on its local input features $\mathbf{x}^{m}$ and locally owned nodes. We term the set of all possible leaf nodes as the \textit{candidate set of party $m$} denoted by $f_k^{m}(\mathbf{x}^m)$.
In this paper, we obtain the following key observation:

\textbf{Key Observation:} Given an input sample $\mathbf{x}$, the prediction value of a federated decision tree model $f_k$ is the weight of the leaf node simultaneously in all candidate sets given by each party.

\begin{figure}[tb]
\centering
\centerline{\includegraphics[width=\columnwidth]{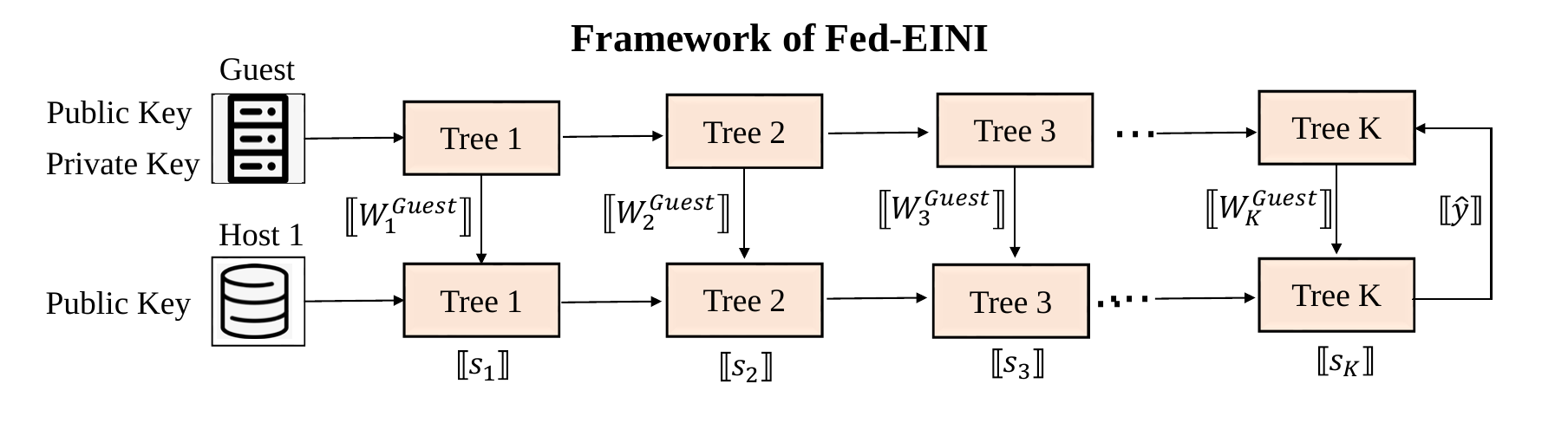}}
\caption{The inference structure of Fed-EINI.}
\label{fig:structure_of_fed-eini}
\end{figure}

The prediction result of a single tree for $\mathbf{x}$ can thus be expressed as the intersection of  results of the sub-models of the trees held by all parties, i.e.,
\begin{equation}\label{eqa:caculation}
\small
  f_k(\mathbf{x})=w_{(j,k)}, \textit{ where }
  j \in \bigcap_{m=1}^{M} f_k^{m}(\mathbf{x}^m).
\end{equation}
\begin{prop}
For the $Tree_k$, the weight $w'_{(j,k)}$ obtained by taking the intersection of results of the sub-models is unique and equal to the weight $w_{(j,k)}$ obtained by MI inference. That means,
\begin{equation}
\small
w_{(j,k)}=f_k(\mathbf{x})=\bigcap_{m=1}^{M} f_k^{m}(\mathbf{x}^m)=w'_{(j,k)}.
\end{equation}
\end{prop}
\begin{proof}
%See Appendix \ref{appendix:a}.
Firstly, every leaf weight $w_{(j,k)}$ corresponds to a path on the $k$-th tree $f_k$. If a sample $\mathbf{x}$ satisfies the splitting condition of the path $w_{(j,k)}$, since  $f_k=\bigcap\limits_{m=1}^M f_k^{m}(\mathbf{x}^{m})$, then the x satisfies the condition of  the path $w_{(j,k)}$ on tree $f_k^m$. And $w_{(j,k)}$ will exist in all the sets $\{f_k^{m}(\mathbf{x}^m)\}_{m=1}^M$, then $f_k(\mathbf{x})$ belongs to the intersection of items $\bigcap\limits_{m=1}^{M} f_k^{m}(\mathbf{x}^m)$ . Secondly, if the intersection is not unique, there will be another weight $w_{(j',k)}$ ($j' \neq j$), and x satisfy all the splitting conditions of the search tree $\{f_k^m\}_{m=1}^M$ on path $w_{(j',k)}$ . Then, $x$ satisfy the splitting condition on path $w_{(j',k)}$ . However, the left and right path of each node of the decision tree are mutually exclusive conditions, so there will not be a sample $\mathbf{x}$ that satisfies two different search paths on a tree at the same time.
\end{proof}
We can thus reformulate the inference result for federated decision tree ensemble model as
\begin{equation}
\small
\hat{y}=\mathcal{G}\Big(\{f_k(\mathbf{x})\}_{k=1}^K\Big)=\mathcal{G}\Big(\{w_{(j,k)}\}_{k=1}^K\Big),
 j\in \bigcap_{m=1}^M f_k^{m}(\mathbf{x}^{m}).
\end{equation}
Based on this key observation, as illustrated in Figure \ref{fig:structure_of_fed-eini}, we propose to compute the candidate sets of each party $f_k^m$ locally and securely compute the inference results with homomorphic encryption (HE) to protect decision paths in each tree model. Since the aggregation functions $\mathcal{G}$ for both GBDT and random forest are linear, we adopt additively HE to improve the computation efficiency instead of fully HE. Therefore, the proposed inference framework is efficient and interpretable, allowing the disclosure of feature meanings of Host parties to Guest. It consists of two stages: parallel calculation stage and synchronization stage. To present the framework concisely, we take a two-party model case as a representative, in which case there are one Guest party and one Host party. It can be easily extended to a multi-party case, which will be demonstrated in Section \ref{subsec:multiparty}.

\subsubsection{Stage 1: Parallel Calculation}
In this stage, each participant generates candidate sets of leaf nodes based on the split conditions of local nodes and local data features, rather than adopting a sequential communication structure as the MI inference approach. The candidate sets are then encoded as a decision vector for synchronization. Specifically, the encoded decision vector for each party is computed as follows.
\begin{itemize}
\item \textbf{Guest}: The encoded vector has as many entries as the number of leaf nodes for a tree model, where each entry is the corresponding weight if the leaf node is in the candidate set and 0 otherwise. That is, we compute an encoded decision vector $\mathbf{W}_k^{\sf{Guest}}\in\mathbb{R}^{T_k}$ for the $k$-th decision tree in the Guest party as
\begin{equation}\label{eqa:guest1}
\small
  \mathbf{W_k}^{\sf{Guest}} \!\!=\! \left[\!s_{(j,k)}\!:\! s_{(j,k)}\!=\!\!\left\{\begin{aligned}
    & w_{(j,k)} && \!if j\!\in\! f_k^{\sf{Guest}}(\mathbf{x}^{\sf{Guest}})\! \\
    & 0 && \!\!otherwise\!
  \end{aligned}\right.  \right],\!
\end{equation}
where $T_k$ is the number of leaf nodes in the $k$-th decision tree model.
\item \textbf{Host}: Similarly, the Host party encodes a candidate set as 1 if the leaf node is in the candidate set and 0 otherwise. The encoded decision vector $W_k^{\sf{Host}}\in\mathbb{R}^{T_k}$ of the $k$-th decision tree for Host as
\begin{equation}\label{eqa:host1}
\small
  \mathbf{W_k}^{\sf{Host}} \!=\! \left[s_{(j,k)}: s_{(j,k)}=\left\{\begin{aligned}
    & 1 && if j\in f_k^{\sf{Host}}(\mathbf{x}^{\sf{Host}}) \\
    & 0 && otherwise
  \end{aligned}\right.  \right].
\end{equation}
%\begin{equation}\label{eqa:host1}
%\small
%  W_k^{\sf{Host}} = \left\{\begin{aligned}
%    & 1 && if j\in f_k^{\sf{Host}}(x^{\sf{Host}}) \\
%    & 0 && otherwise
%  \end{aligned}\right. .
%\end{equation}
\end{itemize}

\subsubsection{Stage 2: Synchronization}
In the synchronization stage, the Guest encrypted weight items as the encoding vector and sent it to the Host to compute the inference result, where the 0's in the encoding vector behave as confusion items to make the decision path indistinguishable for any party. Note that the public key is known to each party while the private key is held only by the Guest.

Specifically, the Guest party first encrypts and synchronizes all encrypted decision vectors $\{\encrypt{\mathbf{W}_k^{\sf{Guest}}}\}_{k=1}^{K}$ to the Host party. Host merges each decision vector of the Guest party with each of its local decision vector and takes the sum of them to obtain the encrypted inference result, i.e.,
\begin{equation}\label{Synchronization}
  \encrypt{\hat{y}}=\sum_{k=1}^{K} \encrypt{S_k}=\sum_{k=1}^{K}\langle \mathbf{1},\encrypt{\mathbf{W}_k^{\sf{Guest}}} \circ \mathbf{W}_k^{\sf{Host}}\rangle.
\end{equation}
Here $\langle \cdot,\cdot\rangle$ denotes the inner product, and $\circ$ denotes the element-wise product of two vectors. Finally, the aggregation value $\encrypt{\hat{y}}$ is sent back to Guest to obtain the inference result through decryption.

The overall procedure of Fed-EINI algorithm is presented in Algorithm \ref{algorithm:fed_eini}.
\SetNlSty{textbf}{}{:}
\IncMargin{1em}
\begin{algorithm}[htb]
\caption{Fed-EINI: an efficient and interpretable inference framework.}
\SetKwInOut{Input}{Input}
\SetKwInOut{Output}{Output}
\Input{$x^{\sf{Guest}},x^{\sf{Host}},\{f_k^{\sf{Guest}}\}_{k=1}^K,\{f_k^{\sf{Host}}\}_{k=1}^K$}
\Output{$Y$}
Set $[q]=0,[S_k]=0$\\
\For{$k=1,\cdots,K$}{
\textbf{Stage 1:} Parallel Calculation\\
Guest$\&$Host: Load parameters of $f_k^{\sf{Guest}}$ or $f_k^{\sf{Host}}$; \\
Guest$\&$Host: Generate $W_k^{\sf{Guest}}$ or $W_k^{\sf{Host}}$ for $x^{\sf{Guest}}$ or $x^{\sf{Host}}$ during tree search according to equation\eqref{eqa:guest1}\eqref{eqa:host1};\\

\textbf{Stage 2:} Synchronization\\
Guest: Encrypt and push $\encrypt{\mathbf{W}_k^{\sf{Guest}}}$ to Host;\\
Host: Pull $\encrypt{\mathbf{W}_{k}}$ from Guest;\\
Host: $\encrypt{S_k}=\langle \mathbf{1},\encrypt{\mathbf{W}_k^{\sf{Guest}}} \circ \mathbf{W}_k^{\sf{Host}}\rangle$;}
Host: Push value $\encrypt{\hat{y}}=\sum\limits_{k=1}^K\encrypt{S_k}$ to Guest;

Guest: Decrypt and get the prediction $\hat{y}$;\\
\label{algorithm:fed_eini}
\end{algorithm}
\begin{figure}[tb]
\centering
\centerline{\includegraphics[width=\columnwidth]{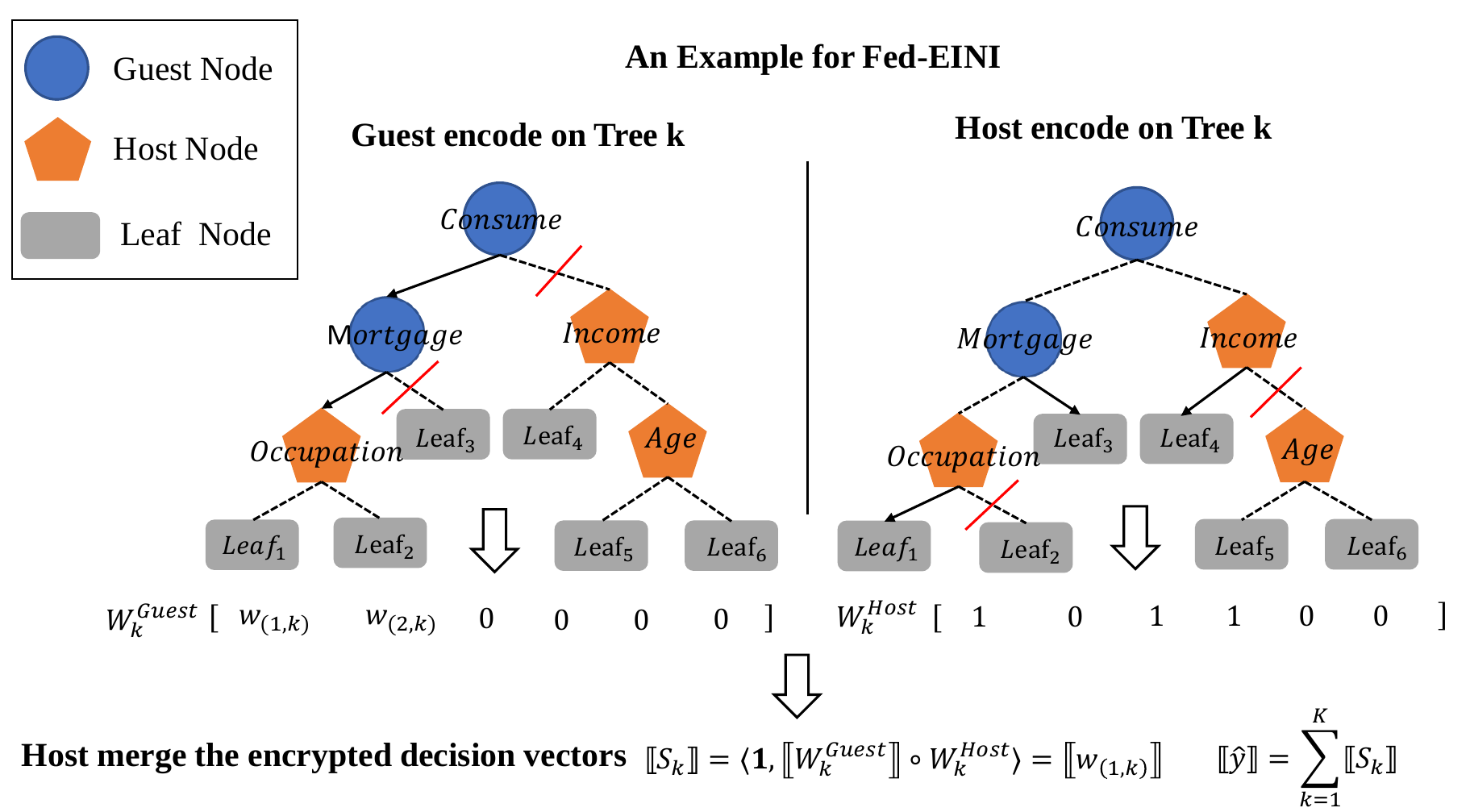}}
\caption{An example for Fed-EINI. }
\label{fig:parallel_calculation_guest}
\end{figure}
Here we present a toy example as illustrated in Figure \ref{fig:parallel_calculation_guest}. In the $k$-th tree model, circles or polygons represent nodes held by Guest or Host. In  parallel calculation stage, for the sample $\mathbf{x}$, it does not satisfy the Guest's split condition of $\textrm{GF}_1$ and $\textrm{GF}_2$. The possible decision path for $\mathbf{x}$ at the Guest is the paths through weight $w_{(1,k)}$ and weight $w_{(2,k)}$.
\begin{equation}\label{eqa:guest2}
\small
  \mathbf{W}_k^{\sf{Guest}} = \big[{w_{(1,k)}}, {w_{(2,k)}}, {0}, {0}, {0}, {0} \big].
\end{equation}
The instance x can't satisfy the Host's split condition of $\textrm{HF}_4$ and $\textrm{HF}_7$. The possible decision path for $\mathbf{x}$ at the Guest is the paths corresponding to $\textrm{Leaf}_1$ , $\textrm{Leaf}_3$ and $\textrm{Leaf}_4$ . These paths are encoded as 1, else 0.
\begin{equation}\label{eqa:host2}
\small
  \mathbf{W}_k^{\sf{Host}} = \big[1 , 0 , 1 , 1 , 0 , 0 \big].
\end{equation}
In synchronization stage of the example, the Guest party encrypts all the entries with HE as $\encrypt{W_k^{Guest}}$and send all encrypted decision vectors $\{\encrypt{\mathbf{W}_k^{\sf{Guest}}}\}_{k=1}^{K}$ to the Host party. The Host party merges each pair of decision vectors as \eqref{Synchronization}. In this example, after Host party merge the decision vectors $\encrypt{W_k^{Guest}}$ and $W_k^{Host}$, we get the encrypt item $\encrypt{w_{(1,k)}}$ as the final result of the k-th tree for x. Finally Host party send $\encrypt{\hat{y}}=\sum_{k=1}^{K} \encrypt{S_k}$ to the Guest party for prediction.

\subsection{Extension To Multi-party Scenarios}\label{subsec:multiparty}
\begin{figure}[tp]
\centering
\centerline{\includegraphics[width=\columnwidth]{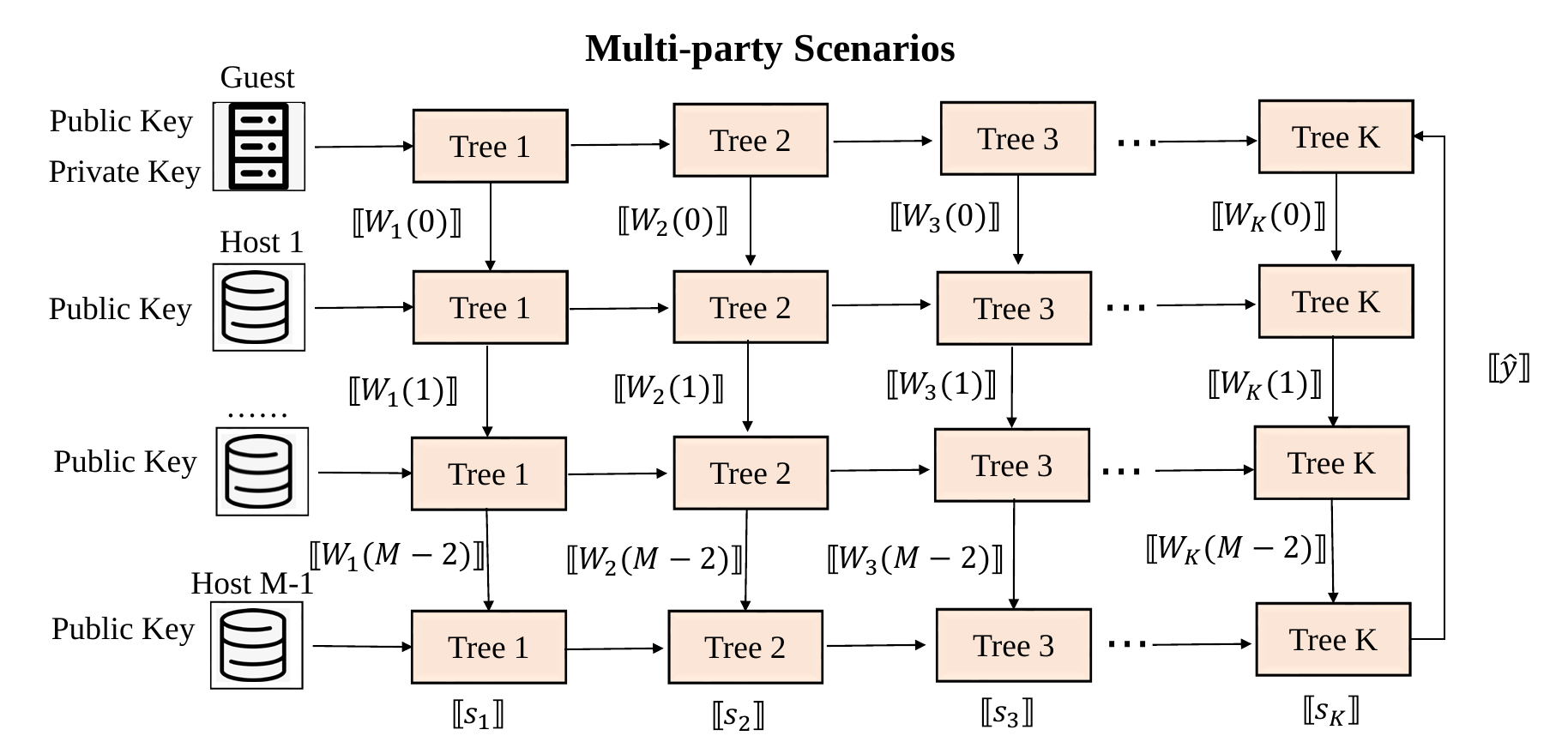}}
\caption{Multi-party security inference.}
\label{fig:Multi-Party security inference}
\end{figure}

The proposed Fed-EINI framework can be easily extended to multi-party scenarios. As shown in Figure \ref{fig:Multi-Party security inference}, the parallel calculation stage is the same, while in the synchronization stage, each Host party updates the encrypted decision vector generated by the Guest and transmits it to the next Host party. Specifically, the $i$-th Host party updates the encrypted decision vector as
\begin{equation}
  \mathbf{W}_k(i) = \mathbf{W}_k(i-1)\circ \mathbf{W}_k^{\sf{Host}_i},~\mathbf{W}_k(0)=\mathbf{W}_k^{\sf{Guest}}
\end{equation}
for $i=0,\cdots,M-1$ with a Guest party (party $0$) and $M-1$ Host parties. The encrypted inference result can be computed as
\begin{equation}
  \encrypt{\hat{y}}=\sum_{k=1}^{K} \encrypt{S_k}=\sum_{k=1}^{K}\langle \mathbf{1},\encrypt{\mathbf{W}_k(M-1)}\rangle.
\end{equation}
The $(M-1)$-th Host party transmits $\encrypt{\hat{y}}$ to the Guest party to decrypt the inference result.

\section{Analysis of Fed-EINI}\label{section:4}
In this section, we provide theoretical analysis on security and interpretability, as well as the efficiency of Fed-EINI.
\subsection{Security and Interpretability}\label{security}
This subsection will demonstrate that Fed-EINI is secure and interpretable while disclosing the meaning of features to the Guest party under the semi-honest and non-collusive assumption. We will first show that the shared information in plaintext across parties will not reveal data privacy, followed by demonstrating that cracking the exchanged ciphertexts across parties is computationally infeasible.

Our method discloses the meaning of features and achieves the same security as the existing framework with the semi-honest assumption. We make a comprehensive comparison on the information available to each party between the proposed Fed-EINI and the multiple-interactive (MI) inference framework adopted in existing frameworks such as SecureBoost \cite{cheng2021SecureBoost}, as illustrated in Table \ref{tab:privacy}.
\begin{table*}[htb]
\caption{Comparison of SecureBoost and Fed-EINI in information sharing}\label{tab:privacy}
\centering
\small
\vskip 0.15in
\begin{tabular}{p{1.4cm}   p{2.8cm}|| p{2.6cm}<{\centering} p{2.6cm}<{\centering}
|| p{2.6cm}<{\centering} p{2.6cm}<{\centering}}
\toprule
  &\multicolumn{1}{c}{}        & \multicolumn{2}{c}{SecureBoost}                                                                                                    & \multicolumn{2}{c}{Fed-EINI}                                                                                            \\\midrule
&           & \makecell[c]{Guest}                                       &\makecell[c]{Host} & \makecell[c]{Guest}                                     & \makecell[c]{Host}                                                                 \\\midrule
\multirow{4}{*}{\shortstack{Model\\Information}} & Model   structure      & $\surd$                             & $\surd$ & $\surd$ & $\surd$ \\& Weights of leaf nodes &
   $\surd$    &    $\times$      &  $\surd$        &   $\times$  \\ & Model parameters
                             & $\surd$     & $\surd$  & $\surd$   & $\surd$
\\  & Splitting rules
                             & Local nodes       & Local nodes                          & Local nodes     & Local nodes \\\midrule
\multirow{6}{*}{\shortstack{Data\\Information}} & Local data & $\surd$ &  $\surd$ &  $\surd$ &  $\surd$ \\
 & Number of features
                             &   $\surd$         & $\surd$      & $\surd$    & $\surd$ \\
                             &
 Meaning of features  &  Local features                                 & Local features                                                        & All features              & Local features       \\  &\multirow{2}{*}{Decision path}
       & Complete decision path & Decision path based on local nodes & Decision path based on local nodes & Decision path based on local nodes \\  \bottomrule
\end{tabular}
\end{table*}
Note that each party is honest-but-curious and the Guest does not collude with any Host given the semi-honest and non-collusive assumption. As shown in Table \ref{tab:privacy}, we categorize the related information used in inference into two classes, i.e., model information and data information. Compared with the MI inference in SecureBoost, the proposed Fed-EINI allows the disclosure of the meanings of features held by passive parties to the Guest while hiding the splitting path information. In the inference process, the disclosure of the splitting path adopted in SecureBoost is privacy-sensitive if the Guest also has the availability to the feature meaning. To address this issue, SecureBoost has proposed to hide the meaning of features held by passive parties, and Guest only knows index numbers of features. Therefore, the Guest knows the split path of each sample but does not know the meaning of the features corresponding to the split path. In the proposed Fed-EINI approach, Guest knows the plain-text feature meaning of each splitting node without any information about the split directions based on other parties' data.
We should notice that the splitting path in Fed-EINI is secured by Paillier cryptosystem \cite{paillier1999cryptosystems}, which is IND-CCA2 secure in the random oracle model.

\begin{figure}[tb]
  \centering
    \centerline{\includegraphics[width=\columnwidth]{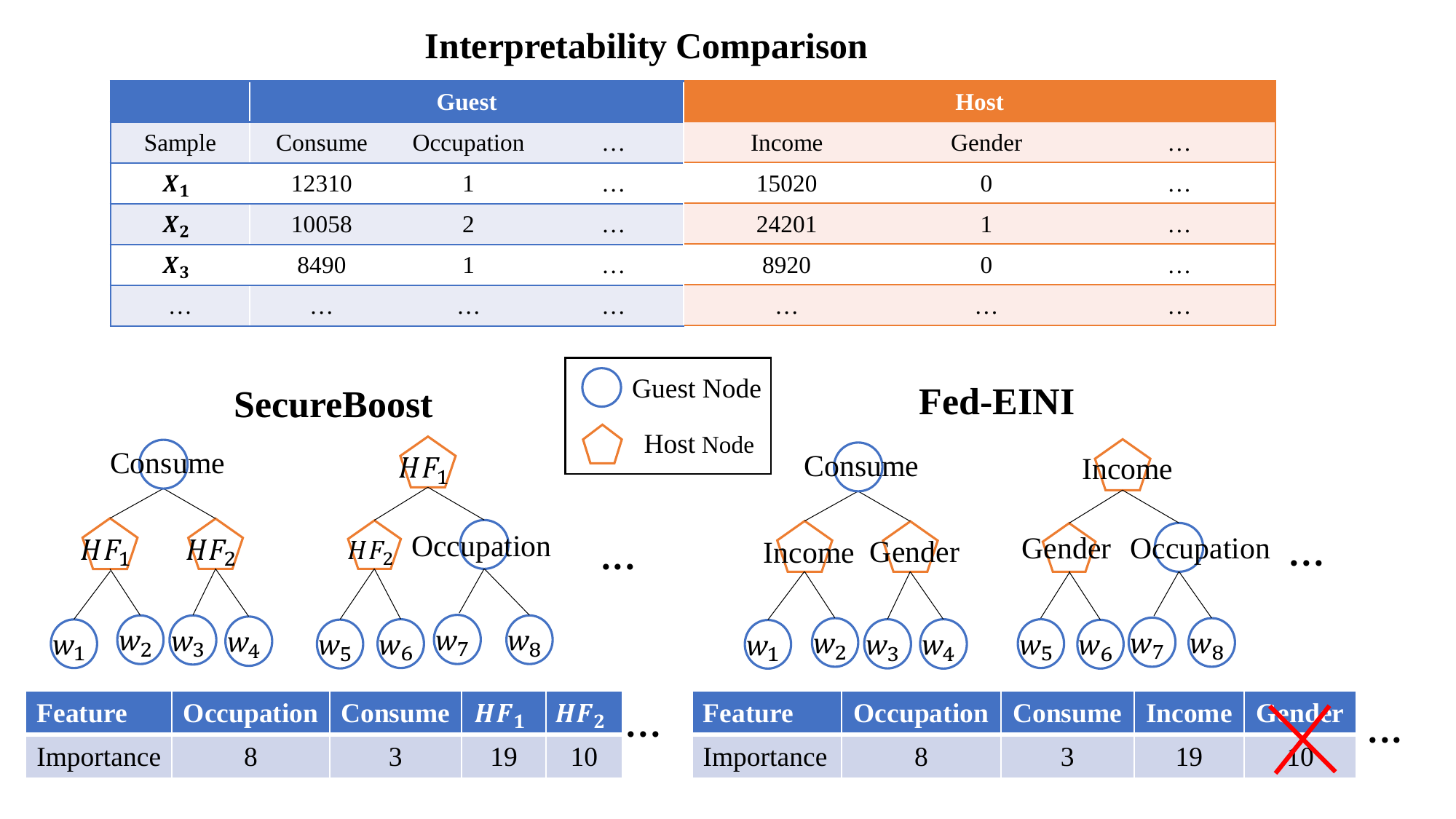}}
    \caption{Comparison of model structure information and feature importance.}
    \label{fig:logo}
\end{figure}
We provide an illustrative example in Figure \ref{fig:logo} to show the differences between Fed-EINI and SecureBoost for GBDT model inference on available model structure information. It can be seen that Fed-EINI provides both the meanings of features and their importance, which is more interpretable and applicable. In real applications,  features held by a party could be already known by other parties due to business dealings in the past. In this case, the approach of anonymizing features in MI inference can still lead to data breaches, while Fed-EINI provides better protection for data privacy. Therefore, Fed-EINI is a more secure and interpretable framework for business use.
\subsection{Efficiency}
Fed-EINI adopts efficient additively homomorphic encryption methods to secure decision paths. Nevertheless, the encryption operations and ciphertext exchanges among parties are still more costly in computation and communication than plaintext adopted in MI inference. We further adopt parallel calculation to improve the computation efficiency and reduce the number of interactions to one round to improve the communication efficiency in Fed-EINI.

\textbf{Parallel Inference:}
For the MI inference framework, the split direction of a node is decided only after its parent node has been visited. With this sequential structure, each party has to wait for the completion of another party to perform local inference. Our proposed inference algorithm decouples the interaction process between multiple parties, and each party generates all the candidates in parallel based on its local splitting condition and local data.

\textbf{One-Round Communication:}
In the MI inference, to predict x, it is necessary to interact at each node level to query whether the split condition is satisfied.
When inferring the sample x in MI inference, communication must be performed at each level. The communication complexity of MI inference is $O(K*Depth)$. Since $f(x)$ can be expressed as the intersection of all participants' candidates $\{f_k^{m}(x^{m})\}_{k=1}^K$, in our framework, the inference of each tree only needs to communicate once at last layer. The total time of communication is $O(K)$.

\section{Experiment and Results}\label{section:5}
In this section, we conduct numerical experiments with the proposed Fed-EINI and the MI inference framework (SecureBoost as representative)  to show the accuracy and efficiency of Fed-EINI.
\subsection{Experiments Setup and Metrics}
Our experiments are conducted on three datasets, Credit1 \cite{dataset:givecredit}, Credit2 \cite{dataset:defaultcredit}, and JDT, to verify the performance of Fed-EINI on classification task. Details of dataset statistics are shown in Table \ref{tab:freq}. The summary of the three datasets is as follows:
\begin{itemize}
    \item \textbf{Credit 1:}It's a open-access credit dataset used to predict whether a user would repay on time. It consists of 30000 data instances, and each instance has 25 attributes;
	\item  \textbf{Credit 2:} It's also a open-access credit scoring dataset to classify whether a user suffers from a financial problem. It consists of 150000 data instances, and each instance has 10 attributes;
	\item  \textbf{JDT}: It's a non-public dataset of JD Technology used for credit scoring. It consists of 512082 data instances, and each instance has 113 attributes.
\end{itemize}
% \begin{enumerate}
%     \renewcommand{\labelenumi}{(\theenumi)}
% 	\item Credit 1: It's a credit scoring dataset used to predict whether a user would repay on time. It consists of 30000 data instances, and each instance has 25 attributes;
% 	\item Credit 2: It's also a credit scoring dataset to classify whether a user suffers from a financial problem. It consists of 150000 data instances, and each instance has 10 attributes;
% 	\item JDT: It's a non-public dataset of JD Technology used for credit scoring. It consists of 512082 data instances, and each instance has 113 attributes.
% \end{enumerate}
%\vspace{-1.4em}
\begin{table}[tbp]
\caption{Description of datasets}
\label{tab:freq}
\begin{center}
\begin{tabular}{c c c c c}
\toprule
Datasets & Instances & Data Size & Attributes \\
\midrule
Credit 1 & 30000   & 5.3M        & 25                   \\
Credit 2 & 150000  & 7.21M          & 10                 \\
JDT     & 512082    & 147M         & 113               \\
\bottomrule
\end{tabular}
\label{tab1}
\end{center}
\end{table}

% \begin{table}[h]
%   \vspace{-0.1em}
%   \caption{Description of datasets}
%   \label{tab:freq}
% \vskip 0.15in
% \centering
% % \small
% \begin{tabular}{c c c c c}
% \toprule
% Datasets & Instances &Attributes & Data Size \\
% \midrule
% Credit 1 & 30000        & 25                 & 5.3M      \\
% Credit 2 & 150000       & 10                 & 7.21M     \\
% JDT     & 512082       & 113                & 147M      \\
% \bottomrule
% \end{tabular}
% % \vskip -0.2in
% \end{table}
To formulate the datasets for each party, we split each dataset into two parts vertically. We randomly select 60\% of the data as the training set, 20\% as the validation set, and the remaining as the test set. To investigate the performance of our proposed method on different dataset scales, we also take 20\%, 80\%, and 100\% of every dataset as subsets.

The Paillier encryption scheme is taken as our additively homomorphic scheme with a key size of 512 bits. All experiments are conducted on two machines with 128GB RAM and 32 CPU cores. We set the maximum number of modeling decision trees as 100 and the maximum tree depth as 4. The experimental party is divided into two sub-sections: accuracy and efficiency.

In this paper, we select AUC (area under the curve) and KS (maximum difference of TPR and FPR) to evaluate the accuracy of our framework. Besides, we use the time cost of the entire inference process to evaluate the efficiency of inference.
\subsection{Security}
    \textbf{Security of the decision path:}The encrypted path is secure, and the local decision path will not be guessed by other parties. In the calculation process shown in Algorithm \ref{algorithm:fed_eini}, the information received by the Host is the encrypted vectors with the same length as each tree model' leaf nodes, and the Host party cannot infer the direction of the split of this sample on the Guest because of the entries encrypted by HE. Besides, what the Guest party receives is the encryption of the final predicted score. It is impossible to infer the direction of the sample on the Host party.

    \textbf{Security of the data privacy:} Since the path is encrypted, disclosing the meaning of features will not reveal the inference data.  As analyzed in \ref{security}, the model information shared between the Guest party and Host party is the same as SecureBoost. The difference in data information available to each party is that the Guest in Fed-EINI knows all the meaning of features of the splitting node, and the Guest in SecureBoost knows the complete decision. The sample decision path held by the Host will not be known by the Guest party; therefore, the  data privacy of the sample will not be revealed.

    However, because the Host party has disclosed the feature name to the Guest, the Guest can know the meaning of each node's feature, which could be regarded as an information leakage of the model. However, in order to make the model user meet the regulatory requirements of interpretability, we think it is acceptable in this case with this level of information leakage.

\subsection{Accuracy Results}
% \begin{figure*}[ht]
%   \vskip 0.2in
%   \centering
%   \subfloat[]{\includegraphics[width=0.33\textwidth]{result1-eps-converted-to.pdf}\label{a}}
%   \subfloat[]{\includegraphics[width=0.33\textwidth]{result2-eps-converted-to.pdf}\label{b}}
%   \subfloat[]{\includegraphics[width=0.33\textwidth]{result3-eps-converted-to.pdf}\label{c}}
%   \caption{Time Cost(s) comparisons between Fed-EINI and SecureBoost. a) Credit 1 dataset. b) Credit 2 dataset. c) JDT dataset.}
%    \label{fig:teaser}
%   \vskip -0.1in
%   \end{figure*}
  \begin{figure*}[ht]
  \centering
  \subfloat[]{\includegraphics[width=0.33\textwidth,height=40mm]{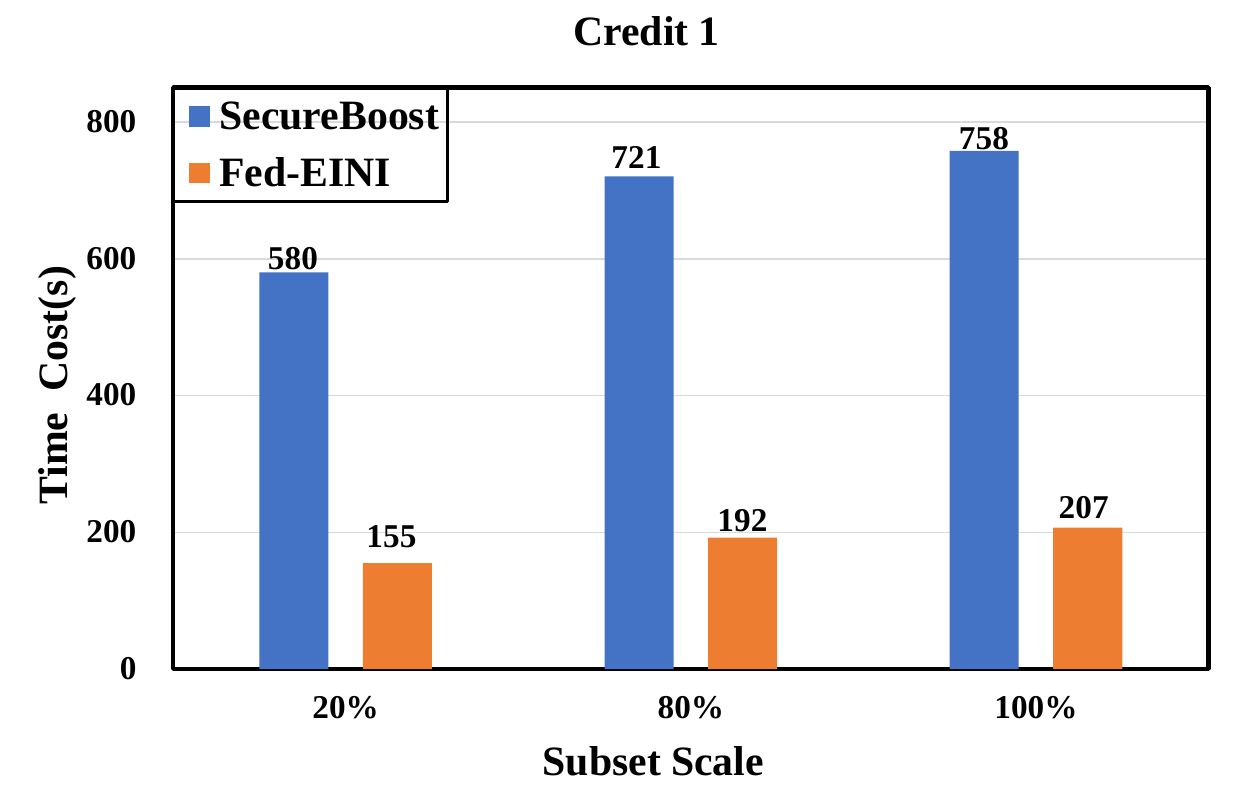}\label{a}}
  \subfloat[]{\includegraphics[width=0.33\textwidth,height=40mm]{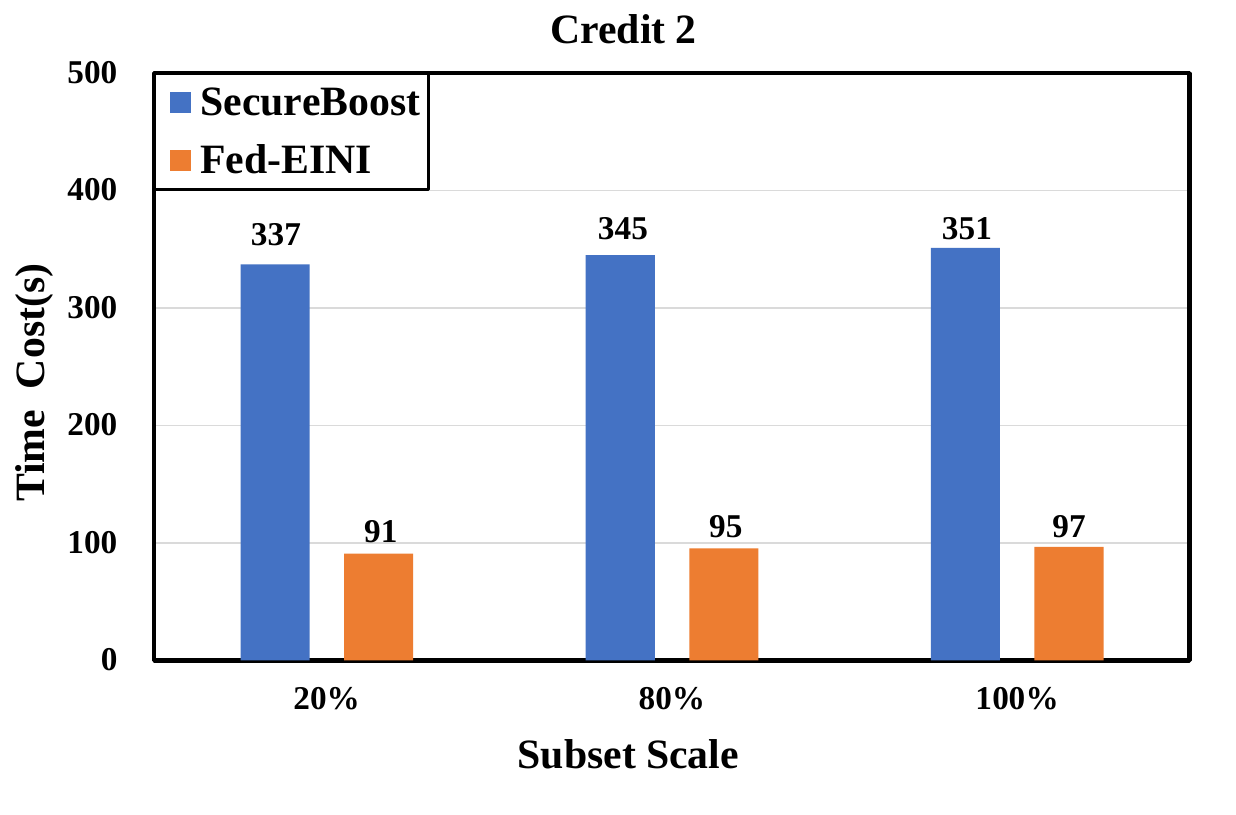}\label{b}}
  \subfloat[]{\includegraphics[width=0.33\textwidth,height=40mm]{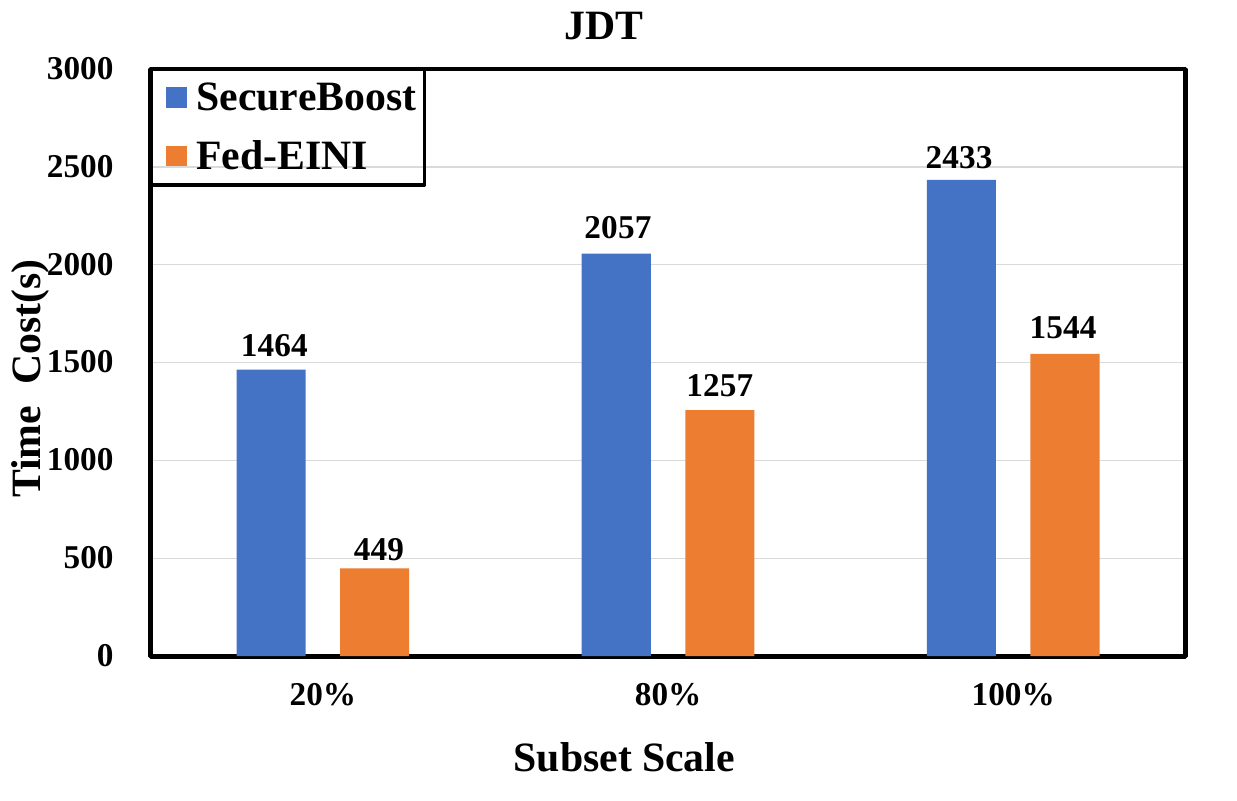}\label{c}}
  \caption{Efficiency comparisons between Fed-EINI and SecureBoost.\\ a) Credit 1 dataset. b) Credit 2 dataset. c) JDT dataset.}
   \label{fig:teaser}
  \end{figure*}
   \begin{figure*}[ht]
  \centering
  \subfloat[]{\includegraphics[width=0.33\textwidth,height=40mm]{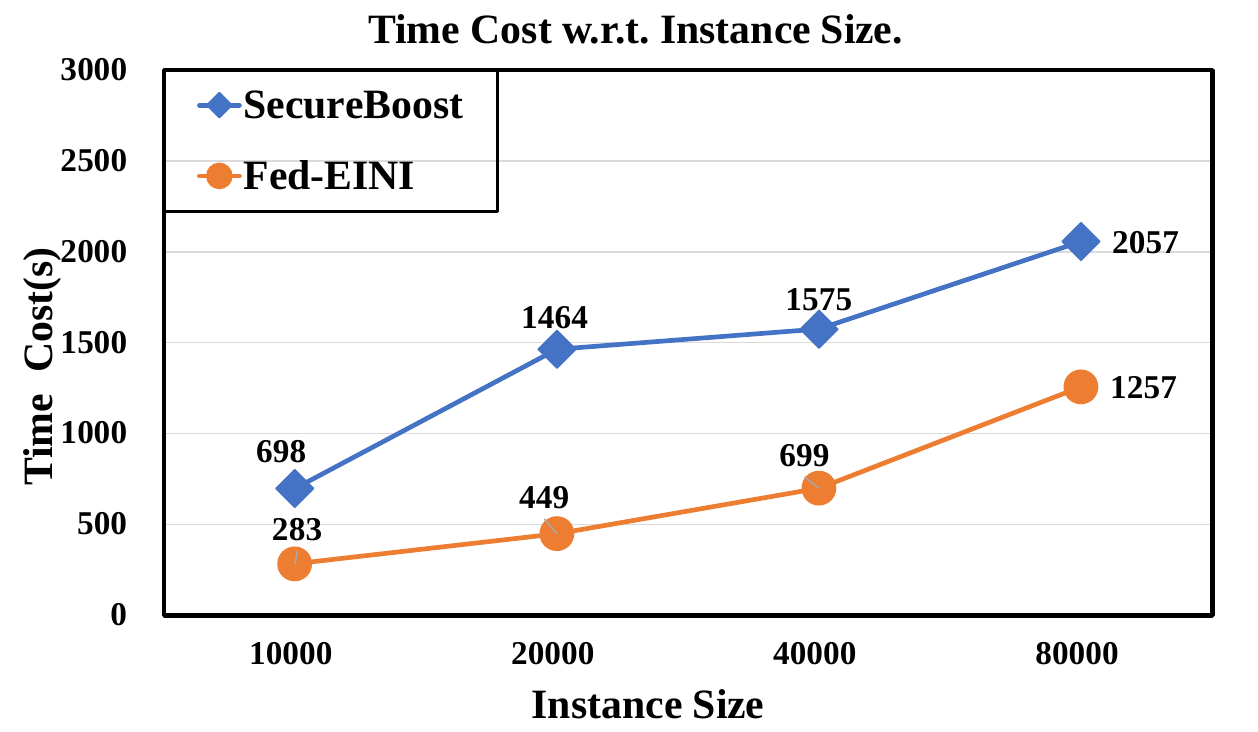}\label{a_1}}
  \subfloat[]{\includegraphics[width=0.33\textwidth,height=40mm]{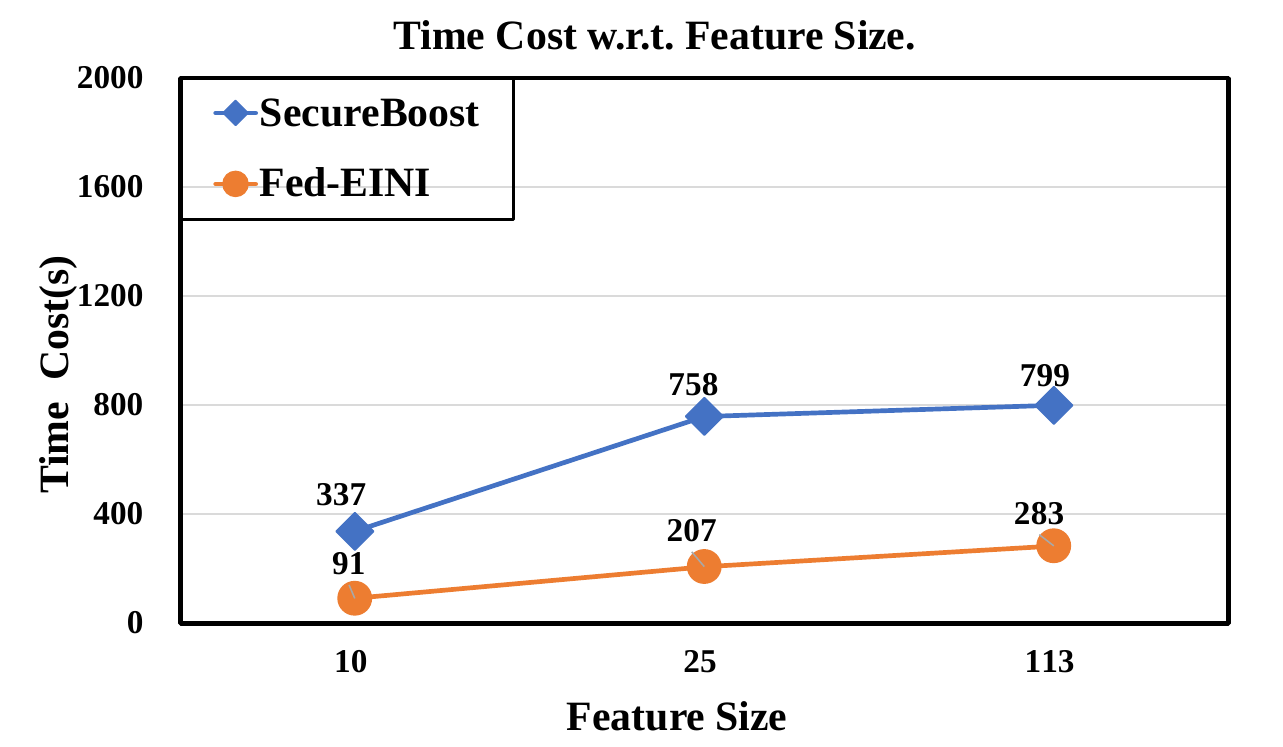}\label{b_1}}
  \subfloat[]{\includegraphics[width=0.33\textwidth,height=40mm]{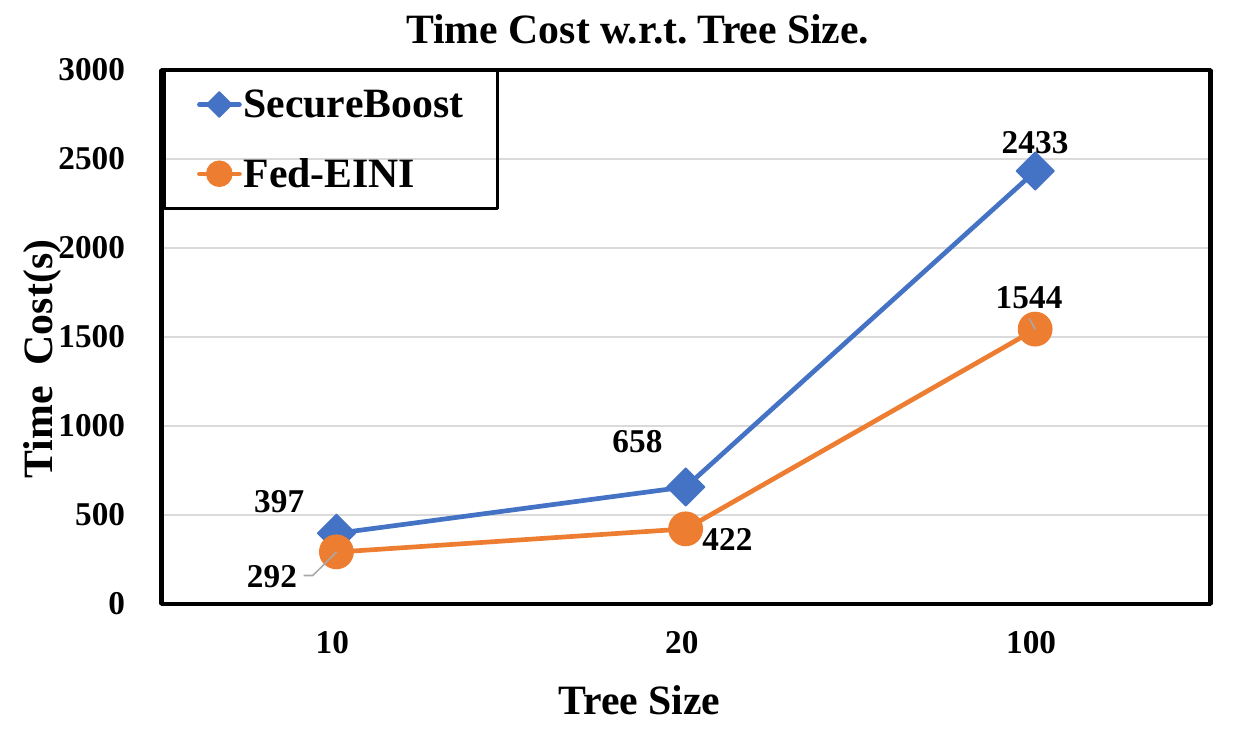}\label{c_1}}
  \caption{Time Cost between Fed-EINI and SecureBoost\\w.r.t  instance size, features size and tree size.}
   \label{fig:teaser}
  \end{figure*}
To prove the accuracy of Fed-EINI, we use the SecureBoost inference method and our improved inference method on the above datasets to conduct experiments and calculate the predictions' evaluation metrics based on the same training model. Table \ref{tab:11} provides the performance comparisons between Fed-EINI and SecureBoost measured by AUC and KS on the randomly sampled subset of each dataset. The metrics of our framework and the MI SecureBoost inference are completely identical to the 3rd decimal digit. Although these algorithms adapt different calculations, both the final result of our inference and the result of MI inference process are the weight of the leaf. Thus the difference between Fed-EINI and SecureBoost is very small. We can infer that our inference framework is lossless with MI inference.

\subsection{Efficiency Results}
\begin{table}[tb]
\caption{Accuracy comparisons between Fed-EINI and SecureBoost}
\label{tab:11}
\begin{center}
\begin{tabular}{c c||c c||c c}
\toprule
\multicolumn{2}{c}{{}} & \multicolumn{2}{c}{SecureBoost}                  & \multicolumn{2}{c}{Fed-EINI}                 \\
\midrule
&Sampling Rate                  &\multirow{1}{*}{\makecell[c]{AUC}} & \multirow{1}{*}{\makecell[c]{KS}} & \multirow{1}{*}{\makecell[c]{AUC}} & \multirow{1}{*}{\makecell[c]{KS}} \\
\midrule
               \multirow{3}{*}{Credit1}    & 20\%   & 0.880                & 60.3                & 0.880                & 60.3                 \\
                            & 80\%   & 0.853                & 55.0                & 0.853                & 55.0                \\
                            & 100\%   & 0.855                & 54.9                & 0.855                & 54.9                \\
\midrule
               \multirow{3}{*}{Credit2}    & 20\%   & 0.773                & 43.0                & 0.773                & 43.0                \\
                            & 80\%   & 0.771                & 41.0                & 0.771                & 41.0                \\
                            & 100\%   & 0.771                & 40.9                & 0.771                & 40.9                \\
\midrule
                \multirow{3}{*}{JDT}      & 20\%   & 0.636                & 19.9                & 0.636                & 19.9                \\
                            & 80\%   & 0.639                & 19.8                & 0.639                & 19.8                \\
                            & 100\%   & 0.638                & 19.7                & 0.638                & 19.7   \\
\bottomrule
\end{tabular}
\label{tab1}
\end{center}
\end{table}
To show the training efficiency of Fed-EINI, we conduct inference experiments on the federated tree algorithm (GBDT).
First, we use the same dataset for model training and then use the SecureBoost inference framework and Fed-EINI framework for inference. We count the time cost of the whole process of inference.

 The Guest party holds five features, and the Host party holds the remaining features. It can be seen from Figure \ref{fig:teaser}, the efficiency of our proposed inference framework method exceeds the SecureBoost inference method. On the credit datasets (Credit1, Credit2), the inference time with our computing method only accounted for about 27\% of the GBDT method. On the JDT data set, the average time consumed by our encoding method only accounts for 49\% of the time consumed by GBDT, and in the worst case, this ratio is only 63.4\%. We guess that this is due to the imbalance in the division of the JDT data set. Guest holds five features, while Host has 108 features. In the parallel calculation stage, the encoding time of the Guest is much shorter than that of the Host. The Host is always in a computing state, so the efficiency improvement is less than Credit datasets.

  Furthermore, we study the effect of the instance size and feature size, and tree size on the efficiency of our proposed framework. We set the maximum tree depth as 4, and vary the instance number in \{10000, 20000, 40000, 80000\} and the feature number in \{10, 25, 113\}, and the tree number in \{10, 20, 100\}. We compare the cost time of the whole inference process to investigate how each variant affects the efficiency. First of all, the Fed-EINI we proposed has significantly less time-consuming than the SecureBoost in each case. In addition, as Figure \ref{a_1} shows, the inference time increases almost linearly with the instance number. There are similar observations on both Figure \ref{b_1} and Figure \ref{c_1}. Besides, we observe that although the number of features increases, the time consumption increase slowly. As the number of features increases, extra features contribute less to the model, so the increasing time consumption for inference is small. Besides, as Figure \ref{c_1} shows, when the tree size is equal to 10, due to the small number of trees, the improvement is not as significant as the case of tree=100. The efficiency of Fed-EINI we proposed will be stable when there are more decision trees.

\section{Conclusion}
In this paper, we studied the inference problem for vertical federated decision tree ensemble models. To address the lack of interpretability, we proposed Fed-EINI to encrypt decision paths to make it secure to disclose feature meanings to the Guest party, namely the model user party. To reduce the high costs resulted from encryption operations and ciphertext exchanges, we proposed an efficient two-stage framework including parallel calculation stage and synchronization stage. It adopted efficient additively homomorphic encryption and parallel calculation to improve computation efficiency and required only one round of information exchange to improve communication efficiency.
Since the decision path is invisible to any party, Fed-EINI provides a more secure and interpretable inference framework than existing works for vertical federated decision tree ensemble models. Experimental results demonstrated the accuracy and efficiency of Fed-EINI compared with state-of-the-art methods.

\bibliographystyle{IEEEtran}

\vspace{12pt}
\color{red}

\end{document}